\documentclass[lettersize,journal]{IEEEtran}
\usepackage{amsmath,amsfonts}
\usepackage{algorithmic}
\usepackage{algorithm}
\usepackage{array}
\usepackage[caption=false,font=normalsize,labelfont=sf,textfont=sf]{subfig}
\usepackage{textcomp}
\usepackage{stfloats}
\usepackage{url}
\usepackage{textcomp}
\usepackage{verbatim}
\usepackage{booktabs}
\usepackage{multirow}
\usepackage{cite}
\usepackage{textcomp}
\usepackage[numbers,sort&compress]{natbib}
\usepackage{graphicx}
\begin{document}

\title{KiGRAS: Kinematic-Driven Generative Model for Realistic Agent Simulation}

\author{
    Jianbo Zhao$^{1,2,*}$, Jiaheng Zhuang$^{2,3,*}$, Qibin Zhou$^{3,*}$, Taiyu Ban$^{1,*}$, Ziyao Xu$^{2,\dagger}$, Hangning Zhou$^{2,\dagger}$, Junhe Wang$^2$, Guoan Wang$^2$, Zhiheng Li$^3$, Bin Li$^1$
    \thanks{$^*$ These authors contributed equally to this work.}
    \thanks{$^\dagger$ Corresponding authors: Ziyao Xu (ziyao.xu@mach-drive.com), Hangning Zhou (hangning.zhou@mach-drive.com).}
    \thanks{$^1$ Jianbo Zhao, Taiyu Ban, and Bin Li are with University of Science and Technology of China, 96 Jinzhai Rd, Hefei 230026, China.}
    \thanks{$^2$ Jianbo Zhao, Jiaheng Zhuang, Ziyao Xu, Hangning Zhou, Junhe Wang, and Guoan Wang are with Mach Drive, Ruixiang Road 88, Wuhu, China.}
    \thanks{$^3$ Jiaheng Zhuang and Zhiheng Li are with the Tsinghua Shenzhen International Graduate School, Tsinghua University, 30 Shuangqing Rd, Haidian District, Beijing 100084, China.}
}

\maketitle

\begin{abstract}
Trajectory generation is a pivotal task in autonomous driving. Recent studies have introduced the autoregressive paradigm, leveraging the state transition model to approximate future trajectory distributions. This paradigm closely mirrors the real-world trajectory generation process and has achieved notable success. However, its potential is limited by the ineffective representation of realistic trajectories within the redundant state space.
To address this limitation, we propose the Kinematic-Driven Generative Model for Realistic Agent Simulation (KiGRAS). Instead of modeling in the state space, KiGRAS factorizes the driving scene into action probability distributions at each time step, providing a compact space to represent realistic driving patterns. By establishing physical causality from actions (cause) to trajectories (effect) through the kinematic model, KiGRAS eliminates massive redundant trajectories.
All states derived from actions in the cause space are constrained to be physically feasible.
Furthermore, redundant trajectories representing identical action sequences are mapped to the same representation, reflecting their underlying actions.
This approach significantly reduces task complexity and ensures physical feasibility.
KiGRAS achieves state-of-the-art performance in Waymo's SimAgents Challenge, ranking first on the WOMD leaderboard with significantly fewer parameters than other models. The video documentation is available at \url{https://kigras-mach.github.io/KiGRAS/}.

\end{abstract}


\section{Introduction}

\IEEEPARstart{S}{mart} autonomous driving (AD) systems depend significantly on the generation of realistic future trajectories for agents, which is crucial for tasks such as motion prediction \cite{feng2023macformer, habibi2021human, zhou2023qcnext}, motion planning \cite{yang2022learning, evans2023high, cimurs2021goal}, and agent simulation \cite{paden2016survey}.
In the context of trajectory distribution modeling, physical feasibility and interaction fidelity are two essential premises for realistic trajectory generation.
Physical feasibility \cite{liu2024spatiotemporal} ensures that trajectories comply with the fundamental principles of vehicle dynamics and kinematics. Interaction fidelity, on the other hand, ensures realism and consistency in modeling the interactions between agents. Adequate capture of these two aspects remains a major challenge in current AD research.

Previous studies \cite{feng2023macformer, zhou2023qcnext}  employ mainly a direct regressive paradigm to approximate future trajectory distributions using Gaussian or Laplace distributions. However, a finite number of these distributions struggle to accurately model complex realistic trajectories, fundamentally limiting the performance of these models. Moreover, the regressive methods inadequately capture agent interactions in future trajectories \cite{seff2023motionlm}, compromising the interaction fidelity of multi-agent trajectories.

Inspired by the success of autoregressive language models \cite{achiam2023gpt}, recent studies have proposed an autoregressive paradigm for trajectory generation \cite{philion2023trajeglish, seff2023motionlm}. 
Unlike previous direct regression in a long future period, these methods model agent and road data as a sequence of short-period state tokens, and process autoregression on these tokens, predicting a short future of trajectories at each step.
This autoregressive paradigm better models the complexity of trajectory distributions and more accurately captures the interactions between agents.

\begin{figure}[t]
\centering
\includegraphics[scale=0.35]{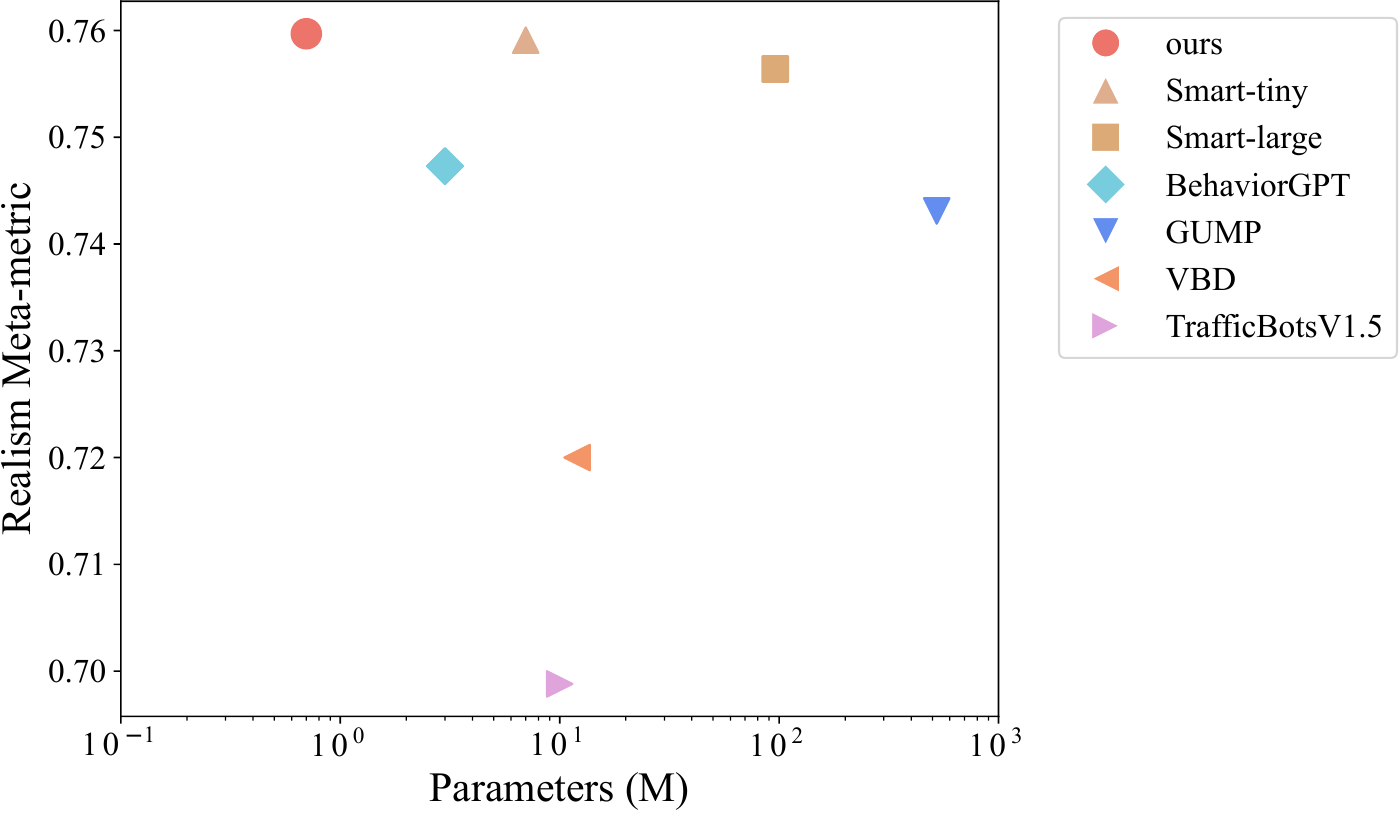}
\caption{Performance comparison of different models based on Parameters (M) and Realism Meta-metric, the overall metric of Waymo's SimAgent. Each marker represents a specific model with its respective parameter size and realism meta score. For more details, see Section \ref{sec:performance_cmp}.}
\label{fig:modelparams}
\end{figure}

However, existing autoregressive methods model directly in the trajectory state space, which is limited by two types of redundancies \cite{sun2024controlmtr}.
First, realistic trajectories constitute only a small subset of the predictive space, and there is a significant presence of trajectories that violate physical laws.
This makes the state space \textit{inefficient} for representing physically feasible trajectories, increasing training difficulty and challenging the physical feasibility of trajectory generation.
Second, different agents exhibit similar kinematic patterns. When modeling state transitions in the state space, a significant amount of information entropy is repeatedly used to describe the same kinematic state transition model. This redundant representation adds \textit{unnecessary} complexity, limiting the effectiveness and potential of the autoregressive paradigm.

To address these challenges, we propose \underline{Ki}nematic-driven \underline{G}enerative Model for \underline{R}ealistic \underline{A}gent \underline{S}imulation (KiGRAS).
KiGRAS presents a new autoregressive paradigm by re-formulating the task from trajectory distribution modeling to \textit{action distribution} modeling.
We introduce the kinematic model to inversely infer control actions from trajectories of consecutive time steps, thus establishing physical causality from the control action (cause) to trajectories (effect).
Instead of existing paradigms learning in the \textit{effect space}, KiGRAS predicts future control actions, which learns in the \textit{cause space} that captures a more fundamental aspect of driving patterns.
In the action space, the massive trajectory redundancies are transformed into concise action sequences, creating a compact representation of driving patterns and greatly reducing the complexity of learning realistic driving behaviors.
Moreover, KiGRAS naturally ensures the physical feasibility of trajectories in all time steps by a kinematic-driven forward update mechanism.
Prior to inference, the state of all agents is updated using the previous state and the predicted control action through a forward kinematic calculation.
This kinematic-driven forward update of agent states builds a hard constraint on state transition, making the state of each time step reachable by control.
Our contributions are listed threefold:

\begin{itemize}
    \item We propose KiGRAS, a novel autoregressive paradigm for the trajectory generation task that learns in the control action space, which makes a compact representation of driving patterns and fundamentally reduces the complexity of realistic trajectory generation.
    \item We propose a kinematic-driven approach for inverse inference of control actions and forward update of agent states, building physical causality from control actions to trajectories and naturally ensuring the physical feasibility of each time step of generated trajectories.
    \item We propose a unified network architecture, which unifies the task definition at each time step and unifies scene representation with the same spatial encoder.
    This unification further simplifies the task complexity, enhances agent interaction modeling, and introduces ease of post fine-tuning for more fine-grained customization.
\end{itemize}

KiGRAS (0.7M) achieves state-of-the-art performance in the Sim Agents challenge\footnote{\url{https://waymo.com/open/challenges/2024/sim-agents/}}, ranking \textbf{1st} on the WOMD leaderboard with significantly fewer parameters than other models, as reported in Fig. \ref{fig:modelparams}.
Interestingly, KiGRAS can embed all human preferences present in the data, similar to InstructGPT \cite{achiam2023gpt}, allowing for the customization of different driving habits. We provide a fine-tuning approach for tailoring driving preferences. For more details, see Section \ref{sec:DPO}.

\section{Related work}

\subsection{Regressive Trajectory Generation}
Predominant deep learning (DL) methods are based on the regressive paradigm to model trajectory distributions in a long future period \cite{grigorescu2020survey}.
In the early stage, some studies \cite{feng2023macformer, zhou2023qcnext} directly take historical trajectories as input and future trajectories as labels, and use the regression loss to supervise the training process. 
These methods assumed that the future trajectory space followed a single Gaussian distribution, which is too poor to model realistic cases.
Subsequently, some studies \cite{zhang2024sparsead, zhuang2024streammotp} generated multimodal trajectories and employed probabilistic regression models to capture the trajectory generation process. These researchers assumed that the future trajectory probability space could be approximated by the superposition of multiple Gaussian \cite{feng2023macformer} or Laplace \cite{zhou2023qcnext} distributions. Yet in practice, finite Gaussian or Laplace distributions are still not sufficient enough to describe the complex realistic distributions of trajectories, especially for long-term future trajectories.
Additionally, some researchers \cite{liao2022itgan, weng2021ptp, xu2023context} adopt generative adversarial network (GAN)-based methods that regard trajectory generation as a task of learning the future trajectory distribution.
However, the long-term future trajectory generation is too complex for GAN, indicating limited performance.

\subsection{Autoregressive Trajectory Generation}
Inspired by the success of autoregressive models \cite{brown2020language, tang2023graspgpt} in the language field, a sequential prediction task sharing many similarities to trajectory generation, several studies have used this autoregressive paradigm to model trajectory distributions, reducing the long future prediction to a step-by-step prediction of the short future period \cite{philion2023trajeglish}.
Seff $et$ $al.$ introduce MotionLM \cite{seff2023motionlm}, which tokenizes trajectories into actions, encodes historical environments and predicts the next action using the previous action during the decoding process. However, these approaches do not update the environment in real-time during decoding, limiting performance in complex scenarios.

Following these explorations, Zhou $et$ $al.$ propose BehaviorGPT \cite{zhou2024behaviorgpt}, which predicts potential trajectory points at each time step using regression modeling. After each prediction, it updates the environmental information and re-encodes it for the next prediction.
Additionally, Wu $et$ $al.$ propose SMART \cite{wu2024smart}, which tokenizes continuous trajectory spaces and uses a classification loss to supervise training. At each time step, SMART predicts the probability distribution of the trajectory space in the next moment and selects trajectory segments using some sampling strategies.

A major difference between our KiGRAS and these methods is that we fundamentally redefine the task in the autoregressive paradigm of trajectory generation. Instead of modeling trajectory distributions filled with massive redundant instances stemming from the same control action, we directly model the space of control actions. This re-formulation provides a much more compact predictive space of realistic driving patterns, thereby significantly reducing the task complexity.
Additionally, KiGRAS inherently ensures the physical feasibility of generated trajectories at each time step by imposing a hard constraint on control reachability between consecutive steps, a criterion not satisfied by the above methods.

\section{Kinematic-Driven Generative Model}

This section introduces details of our KiGRAS.
We start with the problem definition of traditional trajectory generation methods, and then introduce our re-formulation to control action modeling.
Following this, we illustrate details of the control action representation and the kinematic-driven autoregression paradigm.
Finally, we introduce a fine-tuning approach in post-training for driving habit customization.

\begin{figure*}[t]
\centering
\includegraphics[width=0.85\textwidth]{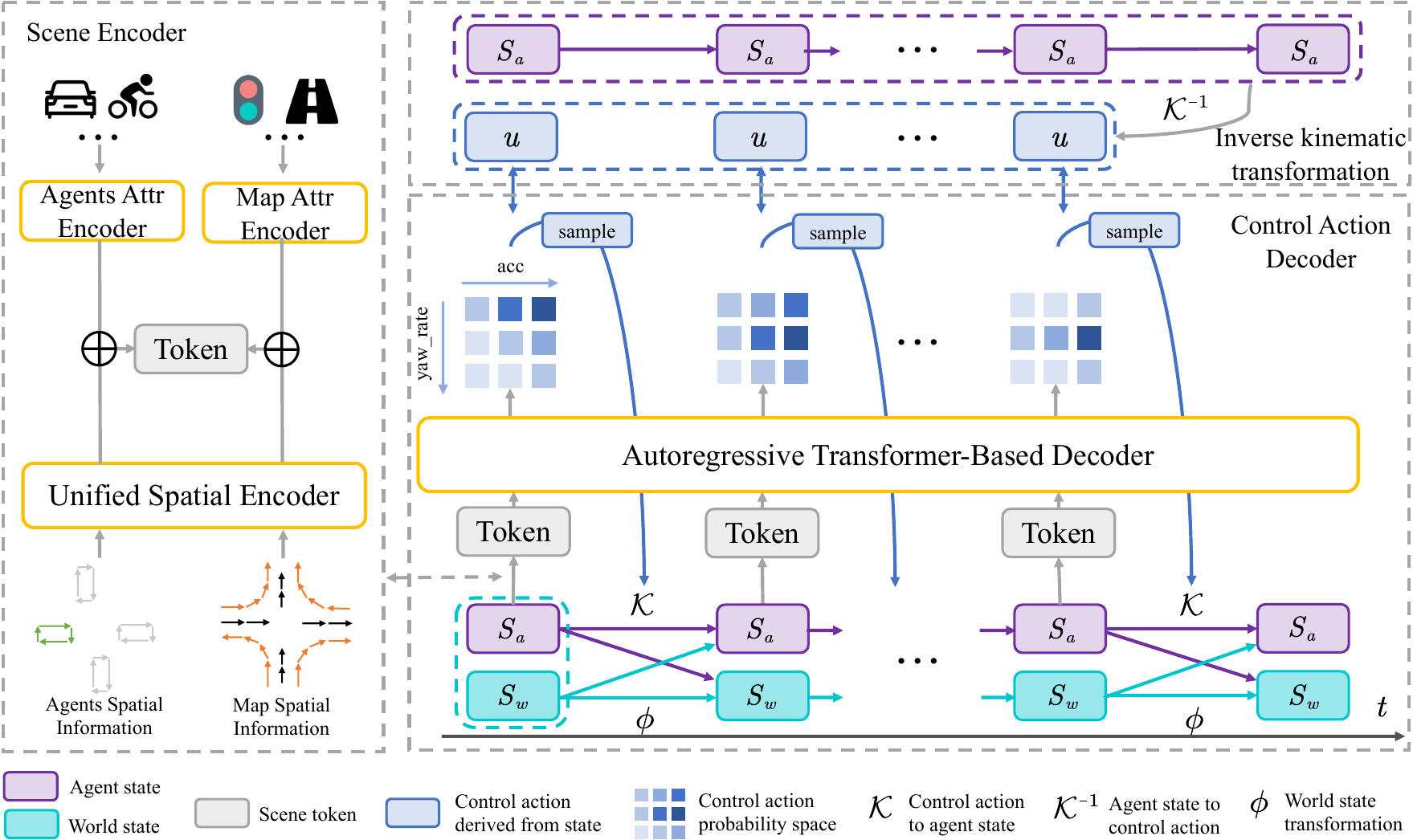}
\caption{The architecture of the KiGRAS framework. First, we solve for the sequence of control actions from trajectory states using the inverse kinematic transformation module. These actions are then represented in a discrete space for processing the forward update of the inference process. To encode the traffic scene, we use a unified spatial encoder to embed the spatial information of all agents and map lines, along with two attribute encoders to describe their traffic roles. Building on this, an autoregressive transform-based decoder is designed to decode the action probability distributions for each state.}
\vspace{-10pt}
\label{fig:framework}
\end{figure*}

\subsection{Problem Definition}

Given the historical states from time $-n$ to time 0 (current time) of one or multiple target agents, $\{\mathcal{S}^\text{a}_{-n}, \ldots, \mathcal{S}^\text{a}_0\}$, and the surrounding world environment $\{\mathcal{S}^\text{w}_{-n}, \ldots, \mathcal{S}^\text{w}_0\}$, the goal is to generate the future states of the target agent from time \(1\) to \(T\), denoted as $\{\mathcal{S}^\text{a}_1, \mathcal{S}^\text{a}_2, \ldots, \mathcal{S}^\text{a}_T\}$. For brevity, we denote the set of target agent's future states from time 1 to time $T$ as $\mathcal{S}^\text{a}_{1:T}$ and the historical states as $\mathcal{S}^\text{a}_{\leq 0}$.

The target agent, denoted as $\mathcal{S}^\text{a}$, includes parameters such as position, heading, and velocity, while $\mathcal{S}^\text{w}$ represents the world environment state, including map data ($S^\text{map}$) and the states of other agents ($S^\text{oa}$) besides the target agent:

\begin{equation}
\mathcal{S}^\text{w} = \{S^\text{map}, S^\text{oa}\}
\end{equation}

This task is formalized to maximize the likelihood of the target agents' future state distribution given the historical states:

\begin{equation}
\label{eq:conditional_distribution}
\arg \max_{\theta} P_{\theta}(\mathcal{S}^\text{a}_{1:T} \mid \mathcal{S}^\text{w}_{\leq 0}, \mathcal{S}^\text{a}_{\leq 0})
\end{equation}

Typically, the future time horizon \( T \) spans several seconds, which is relatively long in the context of a driving scenario. Consequently, this task entails a long-term temporal prediction based on historical data.

\subsection{Re-Formulation to Control Action Modeling}
\noindent\textbf{Probability Factorization.} \quad
The trajectory distribution formulation in Eq. \eqref{eq:conditional_distribution} can be decomposed into a product of conditional distributions over each time step. 

\begin{equation}
\label{eq:trans_invariance}
P(\mathcal{S}^\text{a}_{1:T} \mid \mathcal{S}^\text{w}_{\leq 0}, \mathcal{S}^\text{a}_{\leq 0}) = \prod_{t=0}^{T-1} P(S^\text{a}_{t+1} \mid \mathcal{S}^\text{w}_{\leq t}, \mathcal{S}^\text{a}_{\leq t})
\end{equation}

This decomposition transforms the complex task of long-term trajectory prediction into a series of next-step prediction tasks. Each next-step prediction is significantly less difficult than solving Equation \eqref{eq:conditional_distribution}, as the goal is reduced to predicting the immediate next state based on the history up to that point, rather than tackling the entire sequence at once.

\noindent\textbf{Training Objective Shift} \quad
We further delve into the next-step prediction task and introduce the variable of control action $U$ to enable more fine-grained modeling of the transitions between temporal trajectory states, formalized as:
\begin{equation}
\label{eq:obj_state_next}
\begin{aligned}
   S_{t+1}^\text{a} = \mathcal{K}(S_t^\text{a}, U_t^\text{a}),
\end{aligned}
\end{equation}
Here, $\mathcal{K}$ represents the kinematic bicycle model that is widely acknowledged to describe the state transition given a control action. 
This model governs the transition from the current state $S_t^\text{a}$ to the next $S_{t+1}^\text{a}$ within $U_t^\text{a}$.
Notably, Eq. \eqref{eq:obj_state_next} implies that given \(S^\text{a}_t\) and \(U^\text{a}_t\), the next state \(S^\text{a}_{t+1}\) is independent on any other variables. Based on this, we re-formalize each factor on the right-hand side of Eq. \eqref{eq:trans_invariance} as follows:

\begin{equation}
\begin{aligned}
\label{eq:detailed_prediction}
& P \left(S^\text{a}_{t+1} \mid \mathcal{S}^\text{w}_{\leq t}, \mathcal{S}^\text{a}_{\leq t}\right) \\
= & \int P\left(S^\text{a}_{t+1} \mid \mathcal{S}^\text{w}_{\leq t}, S^\text{a}_{\leq t}, U^\text{a}_t\right) \\
& \quad \times P\left(U^\text{a}_t \mid \mathcal{S}^\text{w}_{\leq t}, \mathcal{S}^\text{a}_{\leq t}\right) \, dU^\text{a}_t \\
= & \int P\left(S^\text{a}_{t+1} \mid S^\text{a}_{t}, U^\text{a}_t\right) \\
& \quad \times P\left(U^\text{a}_t \mid \mathcal{S}^\text{w}_{\leq t}, \mathcal{S}^\text{a}_{\leq t}\right) \, dU^\text{a}_t 
\end{aligned}
\end{equation}
The first equality stems from the law of total probability, and the second equality holds by combining Eq. \eqref{eq:obj_state_next} and the fact that $\{ \mathcal{S}^\text{a}_{t}\}\subseteq \{\mathcal{S}^\text{w}_{\leq t}, \mathcal{S}^\text{a}_{\leq t}\}$.
Furthermore, adhering to the constraint of Eq. \eqref{eq:obj_state_next}, we have that $P\left(S^\text{a}_{t+1} \mid S^\text{a}_{t}, U^\text{a}_t\right) = 1$.
Combining this result with Eqs. \eqref{eq:conditional_distribution}-\eqref{eq:detailed_prediction}, we transform the long-term trajectory state prediction in Eq. \eqref{eq:conditional_distribution} into the next-step prediction task of control actions:
\begin{equation}
\begin{aligned}
\label{eq:task_reformal}
& \underset{\theta}{\arg \max }\, P_{\theta}(\mathcal{S}^\text{a}_{1:T} \mid \mathcal{S}^\text{w}_{\leq 0}, \mathcal{S}^\text{a}_{\leq 0})  \\ 
 \Longleftrightarrow \quad & \underset{\theta^\prime}{\arg \max }\, \prod_{t=0}^{T-1}P_{\theta^\prime}\left(U^\text{a}_t \mid \mathcal{S}^\text{w}_{\leq t}, \mathcal{S}^\text{a}_{\leq t}\right) \\
& \text{subject to} \quad S^\text{a}_{\tau +1} = \mathcal{K}(S_\tau^\text{a}, U_\tau^\text{a}) \\
& t \in \{0,1,\cdots, T-1\},\, \tau\in \{0,1,\cdots,t\}
\end{aligned}
\end{equation}
In summary, our re-formalization\footnote{This formulation assumes the time-invariance hypothesis, which means the probability distribution at each moment can be parameterized using the same set of parameters \cite{benton2020learning}.} is built upon the hard kinematic constraint \( S^\text{a}_{\tau +1} = \mathcal{K}(S_\tau^\text{a}, U_\tau^\text{a}) \), which shifts the training objective from trajectories to control actions. 
Our world transition process ensures temporal interaction consistency among all agents. For readability, we omit the transition of \( S^\text{w} \) here; details can be found in Eq. \eqref{eq:world_state_next}.
This new paradigm reduces the prediction space from complex trajectory distributions to a significantly simpler set of discrete control actions.
Notably, this simplification does not result in any loss of data information. On the contrary, it ensures that all future trajectory states remain physically consistent, thereby improving trajectory modeling and reducing the prediction space.

\subsection{Discrete Representation of Control Action}
\label{sec:control_def}

We represent control actions \( U \) in a discrete space.
Specifically, the probability distribution of a control action \( P(U) \) is defined as the joint probability \( P(A, Y) \), with \( A \) representing acceleration and \( Y \) representing yaw rate. 
To represent realistic control actions, we utilize acceleration rate \( A \) in the range \([-5, 5] \, \text{m/s}^2\) and the yaw rate \( Y \) in the range \([-1.5, 1.5] \, \text{rad/s}\), which generally covers the control actions in the normal driving behavior context.
On this basis, we make a fine-grained split of them into  63 bins.
This results in a discrete action space \({U}\) consisting of totally \(63 \times 63\) control actions, which is flexible enough to model various detailed actions by agents, including vehicles or others.

Compared to previous regressive methods that employ continuous probability distributions, our discrete representation is not dependent on the assumption that probability distribution follows a superposition of multiple prior (Gaussian or Laplace) distributions, thus breaking the limitation of finite prior distributions that are usually too poor to approximate the realistic distributions.

\subsection{Kinematic-Driven Autoregression}
\label{sec:kinematic_gpt}
This section illustrates the details of the kinematic-driven autoregression process in KiGRAS.
We first introduce Inverse Kinematic
Transformation Module, which infers control actions from trajectory data to generate control actions as training labels. Subsequently, we describe the details of our model architecture. Finally, we introduce the training and inference process.

\noindent\textbf{Inverse Kinematic Transformation Module} \quad
To infer control actions between consecutive states, we use a kinematic bicycle model \cite{stone2012kinematic} to describe the physical law between the state \( S \) and the control action \( U \).
The control action $U$ has been defined in Section \ref{sec:control_def}.
For the agent state, it is defined as $s = (x,y,\theta,v)$, including the pose $(x,y,\theta)$ and the velocity $v$.
Note that the definitions of agent states $s$ and control actions $u$, as well as the kinematic model \(\mathcal{K}\), can be substituted with other alternatives that accurately describe the kinematic process.

\begin{figure}[t]
\centering
\includegraphics[scale=0.45]{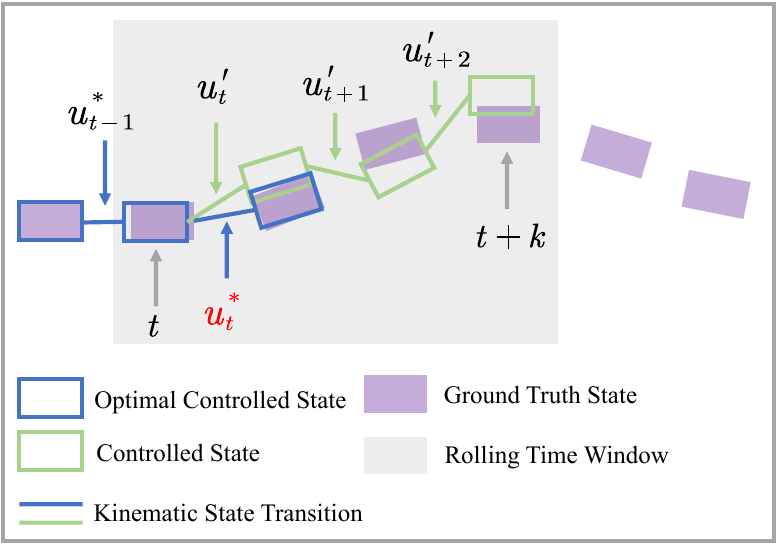}
\caption{
An illustration of the rolling horizon strategy in the inverse kinematic transformation module. We use the \( k \) consecutive states covered by the rolling time window to simultaneously optimize the sequence of control actions for these states (green-lined blocks). The optimal action for the closest future state (blue-lined blocks) is taken as output to mitigate accumulated errors. After these steps, the rolling time window moves forward one step to iteratively solve for all actions.}
\label{fig:tokenization}
\end{figure}

To extract the discrete control action \(U_{0:T-1}\) from the trajectory states \(\mathcal{S}\), we integrate the Model Predictive Controller (MPC) method \cite{garcia1989model} into the transformation process.
Concretely, we use the Rolling Horizon Strategy to solve the discrete control sequence \(\{u_0^*, u_1^*, \ldots, u_{T-1}^*\}\), as shown in Fig. \ref{fig:tokenization}. The process of solving for \(u_t^*\) at each rolling time window can be formulated as follows:

\begin{equation}
\begin{aligned}
    \label{eq:optimization}
    & \min_{u_{t:t+k}}  \sum_{\tau=t}^{t+k} \quad \lVert s_{\tau + 1}^{\text{ctl}} - s_{\tau+1} \rVert \\
    & \text{subject to} \quad s_{\tau + 1}^{\text{ctl}} = \mathcal{K}(s_\tau, u_\tau)
\end{aligned}
\end{equation}
To solve this problem, we use the state \(s_t^{\text{ctl}*}\), obtained by applying \(u_{t-1}^*\) from the previous rolling step, as the initial state for the current rolling time window. Subsequently, we use \(k\) continuous control parameters \(\{u_t^\prime, u_{t+1}^\prime, \ldots, u_{t+k-1}^\prime\}\) in conjunction with the kinematic model \(\mathcal{K}\) to determine the control states \(\{s_{t+1}^{\text{ctl}}, \ldots, s_{t+k}^{\text{ctl}}\}\) within the \(k\)-step time window. We then optimize \(\{u_t^\prime, u_{t+1}^\prime, \ldots, u_{t+k-1}^\prime\}\).

Next, we find the discrete control action in the space \({U}\) that is closest to the optimized continuous control action \(u_t^{\prime*}\) and designate it as \(u_t^*\). Then, we apply the kinematic model \(\mathcal{K}\) to transition the state \(s_t^{\text{ctl}*}\) to \(s_{t+1}^{\text{ctl}*}\) using \(u_t^*\). The resulting state \(s_{t+1}^{\text{ctl}*}\) serves as the initial state for the next rolling step.

\noindent\textbf{Model Architecture} \quad
For the architecture of KiGRAS, we first use a Scene Encoder to encode the multimodal features of agents and environment at each time step. Then, a Control Action Decoder is employed to interact with these features across the temporal sequence and decode the probability distribution of the control action space \( U \) at each time step.

\noindent\textit{Scene Encoder.} \quad At each time step \( t \), we use the same encoder to extract features from the scene information (Fig. \ref{fig:framework} left). This includes the state of the predicted agent, the states of other agents, map information centered around the predicted agent, and traffic light states, all information at each time is encoded into one token.
Notably, we unified spatial representation (USR) of agents and map lines by using identical vectorized features and encoding their spatial information with a Unified Spatial Encoder. For agents, we describe them using four vectors formed by the corner points of their bounding boxes. For map lines, we represent them using vectors between consecutive line points.
On this basis, we embed \(u_{t-1}\) into the current token, referred to as U-Embedding. These operations enhance the model's understanding of the environment and the agent's states. See Section \ref{sec:ablation} for empirical evidence of their effectiveness.

\noindent\textit{Control Action Decoder.} \quad 
Our model utilizes a transformer architecture \cite{vaswani2017attention} with multi-head self-attention (MHSA) to learn complex relationships between agents and maps along the time series and decode control action sequences. We perform self-attention over \(T\) tokens. During training, to prevent information leakage, we leverage causal attention to ensure that all elements can only interact with elements from previous time steps. Additionally, we use teacher forcing \cite{williams1989learning}, which inputs states in driving logs rather than predicted states to ensure consistency in the tasks of each timestep token.

\noindent\textbf{Reactive Inference with World State Update} \quad
To ensure the accurate capture of interactions among all agents, we update the world state of each agent-centric perspective before predicting future control actions.
Formally, we infer the world state at time \( t+1 \), denoted as \( \mathcal{S}_{t+1}^\text{w} \), based on the states of \( N \) agents \(\{S_t^{\text{a}_i}\}_{i=1}^{N}\) and their control actions \( u \) at time \( t \), in addition to the world state at time \( t \), \( \mathcal{S}_t^\text{w} \), as described by the following equation:

\begin{equation}
\mathcal{S}^\text{w}_{t+1} = \phi(\mathcal{S}^\text{map}_t, \{\mathcal{K}(\mathcal{S}^{a_i}_t, u^{a_i}_t)\}_{i=1}^{N})
\label{eq:world_state_next}
\end{equation}
Here, the control action \( u^{{a}}_t \) for all \( N \) agents is sampled from \( P_{\theta}\left(U^{\text{a}}_t \mid \mathcal{S}^{\text{w}}_{\leq t}, \mathcal{S}^{\text{a}}_{\leq t}\right) \), parameterized by our model. This control action is then combined with the current state \( \mathcal{S}^{\text{a}}_t \) and the state transition law \( \mathcal{K} \) to infer \( \mathcal{S}^{\text{a}}_{t+1} \). This process ensures the consistency of the world state for each agent at time \( t+1 \).

In our modeling, we use an agent-centric scene representation for each agent. After inferring all agents' next states, we transform the surrounding environment information of each agent to their agent-centric frame at time \(t+1\). Modeling for each agent allows us to easily fine-tune the model to create drivers with distinct driving habits, which is extremely beneficial for downstream simulation and planning tasks. We demonstrate this flexibility in Section \ref{exp:driving_habit_custom}.

\subsection{Driving Habit Customization}
\label{sec:DPO}

KiGRAS can be further fine-tuned in post-training for better adaptation to practical applications, effectively functioning as a pre-trained model. Various techniques \cite{rafailov2024direct, schulman2017proximal} can be employed to fine-tune this model to accommodate different driving behaviors. Examples of these behaviors include avoiding collisions and driving quickly to meet the requirements of different scenarios. Here, we use the Discriminative Policy Optimization method \cite{rafailov2024direct}.

Specifically, we leverage our pre-trained model to roll out the future trajectories of various prediction targets based on the scene state $x_{i}$ (including map information, agent initialization states, etc.). We then apply expert rules to select winner and losser sample pairs $(y_w^i, y_l^i)$, followed by preference fine-tuning using the Loss function described in Eq. \eqref{eq:dpo}:

\begin{equation}
\label{eq:dpo}
\begin{aligned}
\mathcal{L}_{\mathrm{DPO}}\left(\pi_\theta ; \pi_{\mathrm{ref}}\right) = & -\mathbb{E}_{\left(x, y_w, y_l\right) \sim \mathcal{D}} \left[ \log \sigma \left( \beta \log \frac{\pi_\theta\left(y_w \mid x\right)}{\pi_{\mathrm{ref}}\left(y_w \mid x\right)} \right. \right. \\
& \left. \left. - \beta \log \frac{\pi_\theta\left(y_l \mid x\right)}{\pi_{\mathrm{ref}}\left(y_l \mid x\right)} \right) \right]
\end{aligned}
\end{equation}

The model $\pi_\theta$ represents the policy we aim to optimize, while $\pi_{\mathrm{ref}}$ serves as the reference policy. In our approach, $\pi_{\mathrm{ref}}$ refer to our pre-trained model.  The scaling factor $\beta$ and the sigmoid function $\sigma$ are employed to fine-tune preferences, encouraging the learned policy $\pi_\theta$ to favor trajectories $y_w$ over $y_l$ based on the reference policy $\pi_{\mathrm{ref}}$.

\begin{figure*}[t!]
\centering
\captionsetup[subfloat]{font=small,labelfont=bf,textfont=normalfont}

\subfloat[Multi-agent interaction.\label{fig:subfig1}]{
    \includegraphics[width=0.95\textwidth]{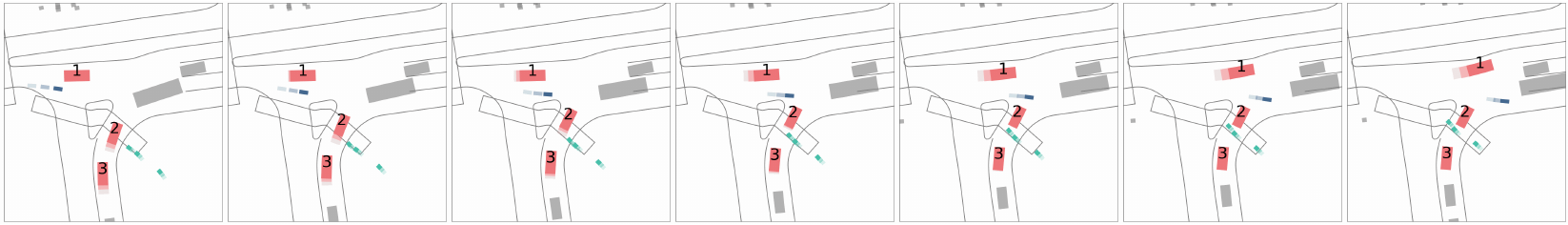}
    \hfill  
}
\vspace{0.1em}

\subfloat[Keep lane.\label{fig:subfig2}]{
    \includegraphics[width=0.95\textwidth]{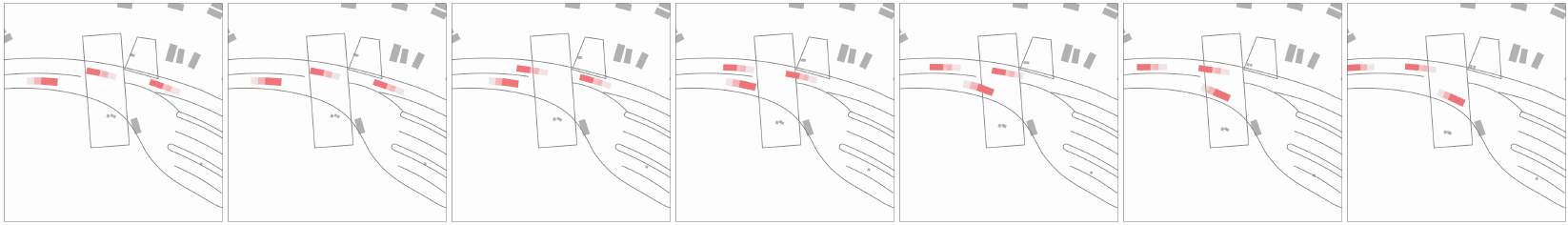}
    \hfill
}
\vspace{0.1em}

\subfloat[Unprotected right turn.\label{fig:subfig3}]{
    \includegraphics[width=0.95\textwidth]{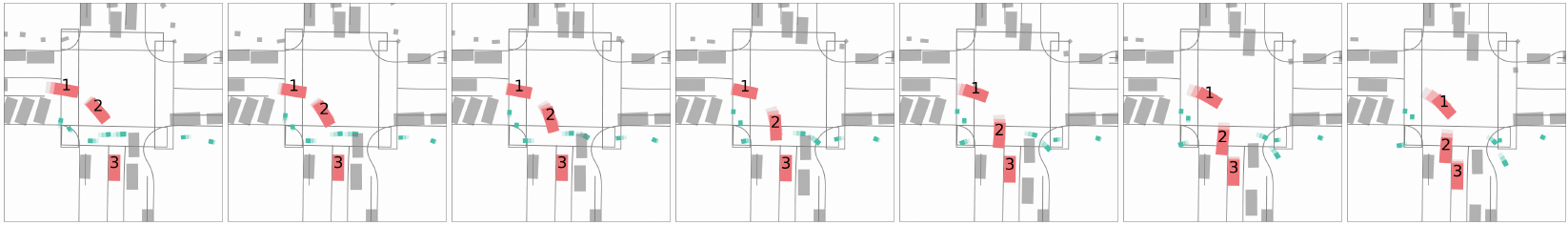}
    \hfill
}
\vspace{0.1em}

\subfloat[Car (index 1) driving against the flow of traffic.\label{fig:subfig4}]{
    \includegraphics[width=0.95\textwidth]{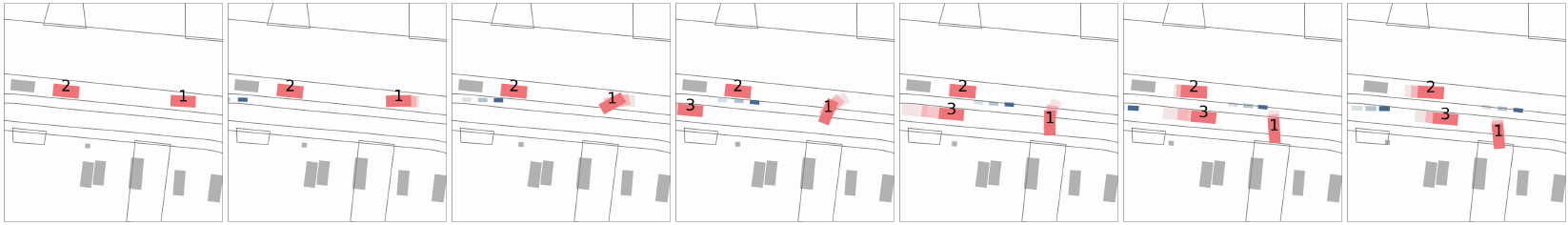}
    \hfill
}

\caption{
Qualitative results of closed-loop simulation. We present four representative scenarios generated by KiGRAS. Trajectories of all agents in these scenarios are generated by KiGRAS. In a fully closed-loop setting, we simulate the future for 8 seconds. Agents of interest are highlighted with distinct colors (some with labels), and their one-second historical trajectories are shown to illustrate their speed changes.}
\label{fig:main}
\end{figure*}

\section{Experiment}
This section presents the experiment results and analysis.
First, we introduce the dataset used and the metrics for subsequent quantitative analysis. Next, we perform a quantitative performance analysis of our method compared to several other modeling approaches. Then, we present closed-loop simulation results of our model in several classic scenarios for qualitative analysis. Following this, we demonstrate the fine-tuning results of specific driving styles based on the pre-trained driver model. Finally, we conduct ablation studies on various modules used during model training.
\subsection{Dataset and Metrics}
We conducted experiments on the Waymo Motion Dataset \cite{sun2020scalability} v1.2, containing 103,354 real human driving scenarios. Our model was trained exclusively on the training set with a 2 Hz frequency.
To ensure a fair comparison with other methods, we used the metrics provided by Waymo SimAgents, which evaluate the trajectories of all agents over an 8-second closed-loop simulation. These metrics include kinematic (Kin.), agent interaction (Inter.), and map-related (Map.) aspects. These metrics assess the speed and acceleration, frequency of non-collision events, distance to leading vehicles, time to collision, and interaction with the road map, respectively. 
REALISM (Real.) is an overall metric derived from a weighted combination of the aforementioned metrics, as explicitly stated by the Waymo SimAgents challenge.
Other metrics include acceleration (Acc), Jerk, and Collision Rate.

\subsection{Implementation Details}
\label{sec:implement_detalis}

We encode each agent and each line into a 64-dimensional vector. For each prediction target, we focus only on the nearest 64 obstacles. We use a 3-layer self-attention mechanism to integrate all scene elements into a 64-dimensional scene token, followed by 3 layers of causal interaction. Regarding DPO, the parameter \(\beta\) in Eq. \eqref{eq:dpo} is set to 1. Details on the selection of positive and negative samples can be found in Section \ref{exp:driving_habit_custom}. During training, we set the batch size to 256 and the learning rate to 2e-4, utilizing the OneCycleLR scheduler to dynamically adjust the learning rate.
We use the Coordinated Turn Rate Acceleration (CTRA) model to describe the transformation $\mathcal{K}$.
During training, we use Cross Entropy Loss to supervise the learning of action probability space.

\subsection{Performance Comparison}
\label{sec:performance_cmp}
Due to changes in Waymo's metrics in 2024 and the closure of submissions to the 2023 leaderboard, we only compared our method with those appearing on the 2024 leaderboard for fairness, including SMART \cite{wu2024smart}, BehaviorGPT \cite{zhou2024behaviorgpt}, GUMP \cite{hu2024solving}, MVTE \cite{wang2023multiverse}, VBD and TrafficBOTv1.5 \cite{zhang2024trafficbots}.
The results are reported in Table \ref{tab:cmp}.
Our model achieved advanced performance on the overall metric of the Waymo SimAgents challenge, surpassing contemporary methods. Additionally, our model has only 0.7M parameters, making it significantly smaller and more lightweight than both 2024 and previous models. We also compared the parameter counts of our model with those of past methods.

\begin{table}[h]
\footnotesize
\setlength{\tabcolsep}{3pt}
\caption{Comparison results of KiGRAS and state-of-the-art approaches in SimAgents Challenge.}
\centering
\begin{tabular}{@{}lccccc@{}}
\toprule
Method         & Real.  & Kin.  & Inter.  & Map.  & minADE \\ \midrule
SMART-96M      & 0.7564              & 0.4769            & 0.7986              & 0.8618            & 1.5501 \\
SMART-7M       & 0.7591              & 0.4759            & 0.8039              & 0.8632            & \textbf{1.4062} \\
BehaviorGPT-3M    & 0.7473              & 0.4333            & 0.7997              & 0.8593            & 1.4147 \\
GUMP-523M         & 0.7431              & \textbf{0.4780}            & 0.7887              & 0.8359            & 1.6041 \\
MVTE           & 0.7302              & 0.4503            & 0.7706              & 0.8381            & 1.6770 \\
VBD-12M            & 0.7200              & 0.4169            & 0.7819              & 0.8137            & 1.4743 \\
TrafficBOTv1.5-10M & 0.6988              & 0.4304            & 0.7114              & 0.8360            & 1.8825 \\ 
KiGRAS-0.7M       & \textbf{0.7597}              & 0.4691            & \textbf{0.8064}              & \textbf{0.8658}            & 1.4383 \\ \bottomrule
\end{tabular}
\label{tab:cmp}
\end{table}

\subsection{Qualitative Results}

In this experiment, we present the results of closed-loop simulations over 8 seconds for all agents using the Pre-trained Driver Model in four classic scenarios, as shown in Fig. \ref{fig:main}. In Fig. \ref{fig:subfig1}, the red vehicle 2 yields to the blue non-motorized vehicle, and the red vehicle 3 decelerates to allow pedestrians to cross. In Fig. \ref{fig:subfig2}, three red vehicles maintain their lanes on a complex-shaped road. In Fig. \ref{fig:subfig3}, red vehicles 1 and 2 perform unprotected right turns and yield to pedestrians, with vehicle 2 attempting an efficient lane change to overtake vehicle 3. In this case, we also observed an issue where pedestrians tend to have close encounters with the edges of motor vehicles.

Additionally, we discovered a particularly interesting case in the Waymo test set where vehicle 1 stopped against the flow of traffic, a scenario rarely seen in the training set (shown as Fig. \ref{fig:subfig4}). Remarkably, our model demonstrated a degree of generalization to this corner case; vehicle 1 promptly made a left turn into a parking space on the left side, allowing other vehicles to proceed smoothly.

\subsection{Results of Driving Habit Customization}
\label{exp:driving_habit_custom}

We fine-tuned our model using the DPO method introduced in Section \ref{sec:DPO}. Based on KiGRAS trained on the Waymo Motion training dataset, denoted as the Pre-trained Driver (PDriver) model, we developed three specialized drivers: the Safety Driver (SDriver), optimized for collision avoidance; the Fast Driver (FDriver), optimized for maximizing speed; and the Comfort Driver (CDriver), optimized for minimizing jerk.
Initially, we sampled 256 potential future trajectories from the Pre-trained Driver Model using a top-p sampling strategy, with data from other agents sourced from logs. For the Safety Driver (SDriver), we selected a non-collision trajectory as the positive sample and a collision trajectory as the negative sample. The construction of positive and negative sample pairs for the Fast Driver (FDriver) differed slightly: we chose the fastest non-collision trajectory as the positive sample, while the negative sample selection mirrored that of the Safety Driver. For the Comfort Driver (CDriver), we selected the trajectory with the lowest maximum jerk that did not result in a collision as the positive sample, with the negative sample selection remaining consistent with the other driver profiles.
We conducted closed-loop simulations to evaluate the performance of each specialized Driver on the Waymo test dataset. Specifically, we used each custom Driver to control the ego-vehicle, while all other agents were controlled by the Pre-trained Driver (PDriver) to ensure a fair comparison. We selected the action with the maximum probability from the model's predicted action space. Results are shown in Table \ref{tab:ablation_dpo}.

\begin{table}[h]
\footnotesize
\centering
\caption{Comparison with different style driver models.}
\setlength{\tabcolsep}{3pt}
\label{tab:ablation_dpo}
\begin{tabular}{@{}c|cccc|ccc@{}}
\toprule
\multirow{2}{*}{Model} & \multirow{2}{*}{\begin{tabular}[c]{@{}c@{}}Speed\\ (m/s)\end{tabular}} & \multirow{2}{*}{\begin{tabular}[c]{@{}c@{}}Acc\\ (m/s$^2$)\end{tabular}} & \multirow{2}{*}{\begin{tabular}[c]{@{}c@{}}Jerk\\ (m/s$^3$)\end{tabular}} & \multirow{2}{*}{\begin{tabular}[c]{@{}c@{}}Jerk$_\text{max}$\\ (m/s$^3$)\end{tabular}} & \multicolumn{3}{c}{Collision Rate (\textperthousand)} \\  
                       &                                                                        &                                                                                                &                                                                                          &                                                                                              & 3s                    & 5s                    & 8s                   \\ \midrule
PDriver                & 5.211                                                                  & 0.402                                                                                          & 0.065                                                                                    & 0.195                                                                                        & 2.118                 & 4.169                 & 7.959                \\
SDriver                & 5.534                                                                  & 0.413                                                                                          & 0.077                                                                                    & 0.231                                                                                        & 1.850                 & 3.389                 & 6.354                \\
CDriver                & 5.017                                                                  & 0.364                                                                                          & 0.051                                                                                    & 0.155                                                                                        & 2.096                 & 4.236                 & 8.271                \\
FDriver                & 5.707                                                                  & 0.440                                                                                          & 0.076                                                                                    & 0.223                                                                                        & 2.185                 & 4.860                 & 9.676                \\ \bottomrule
\end{tabular}
\end{table}

We observed that Safety Driver demonstrated a 1.6052‰ reduction in collision rate over 8 seconds compared to PDriver. Conversely, Fast Driver achieved a 0.4872 m/s increase in average speed; however, this speed increase resulted in a higher collision rate. Additionally, Comfort Driver exhibited a 21.5\% decrease in average jerk and a 20.5\% decrease in maximum jerk.

\begin{figure}[h]
    \centering
    \captionsetup[subfloat]{font=small,labelfont=bf,textfont=normalfont}
    
    \subfloat[Pre-trained Driver.\label{fig:pretrained}]{
        \includegraphics[width=0.45\columnwidth]{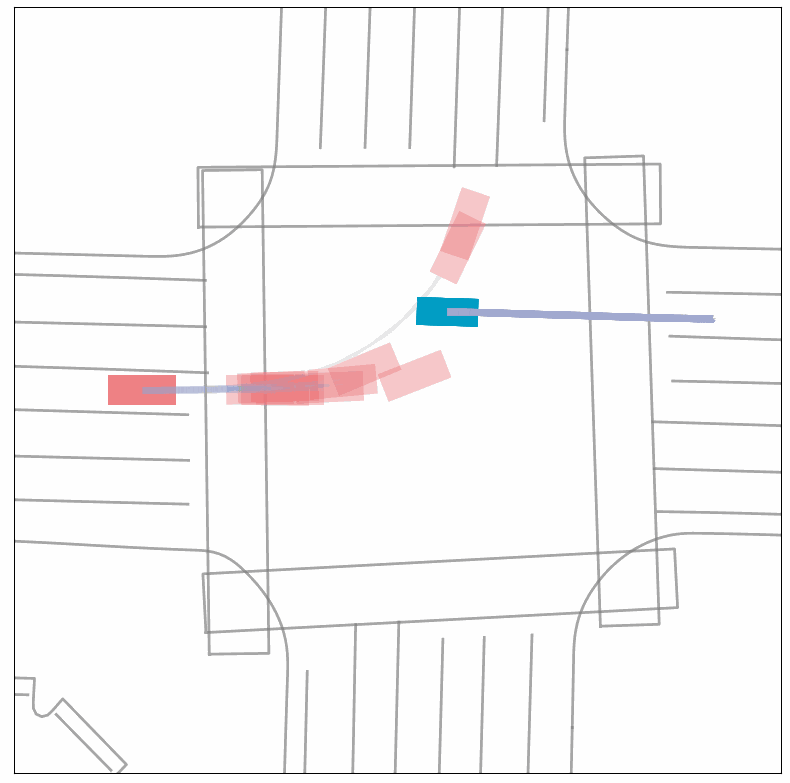}
    }
    \hfill
    \subfloat[ Fast Driver.\label{fig:finetune}]{
        \includegraphics[width=0.45\columnwidth]{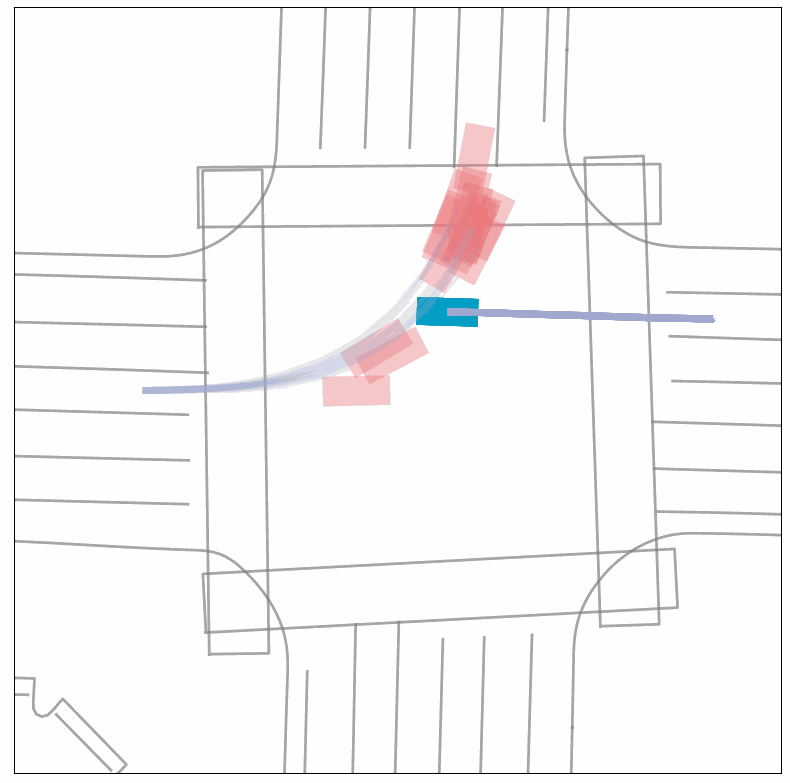}
    }
    
    \caption{Performance Comparison of Pre-trained Driver and Fast Driver Models. Both models were simulated 16 times for 8 seconds each under semi-closed-loop settings.}
    \label{fig:compare}
\end{figure}

To provide a clearer understanding of the changes before and after fine-tuning, we conducted 16 times of semi-closed-loop simulations over 8 seconds. In these simulations, the red ego vehicle was controlled by the model, while the blue other agents were replayed from logs. The results are illustrated in Fig. \ref{fig:compare}, where it is evident that Fast Driver operates at a significantly higher speed.

\subsection{Ablation Study}
\label{sec:ablation}

We conducted the ablation study on three components of our pre-trained model, training on the dataset for 1.5 million iterations.
We reported the cross-entropy (CE) loss on the validation set, with the results shown in Table \ref{tab:ablation_study}.
Introducing "Causal-Attn" in the temporal sequence reduced the CE by 0.087 compared to the base model, indicating better fitting to the data distribution through temporal interactions.
Unified the positions and shape of both agents and map with same representation ("USR") reduced the CE by 0.04, enhancing feature interaction and data distribution learning.
Finally, incorporating previous control action embedding as input ("U-Embedding") further reduced the CE by 0.053, as the model leveraged past control information to improve the learning of the current control action probability space.

\begin{table}[h]
\centering
\caption{Ablation for various modules in Pre-trained Driver Model.}
\begin{tabular}{@{}c|ccc|c@{}}
\toprule
ID & Causal-Attn & USR       & U-Embedding & CE $\downarrow$ \\ \midrule
1  &             &           &           & 1.966                        \\
2  & $\checkmark$   &           &           & 1.879                        \\
3  & $\checkmark$   & $\checkmark$ &           & 1.839                        \\
4  & $\checkmark$   & $\checkmark$ & $\checkmark$ & 1.786                        \\ \bottomrule
\end{tabular}
\label{tab:ablation_study}
\end{table}

\section{Conclusions and Future work}

This paper proposes KiGRAS, a novel autoregressive trajectory generation paradigm that transforms the task to model the distribution of control actions, thereby capturing the physical causality of trajectories. This new task definition offers a much more compact representation of realistic driving patterns, fundamentally reducing task complexity. Additionally, KiGRAS inherently ensures the physical feasibility of generated trajectories through task re-formulation, providing a significant advantage over previous methods.
KiGRAS achieves top performance in the Waymo SimAgents Challenge with a significantly smaller parameter scale, opening a new frontier for the development of next-generation Deep Learning (DL) paradigms in the autonomous driving domain.

For future work and applications, our model has the potential to be used as a simulator trained on real driving data, which opens up opportunities for its application in reinforcement learning tasks. Additionally, our approach can be easily extended to motion prediction and planning tasks. A key focus of our research will be on how to sample actions from the action space to meet specific task requirements effectively.

\bibliographystyle{IEEEtran} 
{\footnotesize 
\bibliography{IEEEexample}
}

\newpage

\end{document}